\documentclass[letterpaper]{article} 
\usepackage{aaai2026}  
\usepackage{times}  
\usepackage{helvet}  
\usepackage{courier}  
\usepackage[hyphens]{url}  
\usepackage{graphicx} 
\usepackage{natbib}  
\usepackage{caption} 
\usepackage{multicol}

\usepackage{amssymb}
\usepackage{amsmath}
\usepackage{algorithm}
\usepackage{algpseudocode}
\usepackage{multirow}
\usepackage{float}

\usepackage{newfloat}
\usepackage{listings}
\DeclareCaptionStyle{ruled}{labelfont=normalfont,labelsep=colon,strut=off} 
\floatstyle{ruled}
\newfloat{listing}{tb}{lst}{}
\floatname{listing}{Listing}
\nocopyright 

\title{AAAI Press Anonymous Submission\\Instructions for Authors Using \LaTeX{}}

\title{SCOUT: Toward Sub-Quadratic Attention via \underline{S}egment \underline{C}ompression for \underline{O}ptimized \underline{U}tility in \underline{T}ransformers}

\definecolor{mypurple}{RGB}{128,0,128}

\author{
    Aref Jafari $^{1, 2}$ \quad
    Yuhe Fan $^{1}$, \quad
    Benyamin Jamialahmadi $^{1}$, \quad
    Parsa Farinneya $^{1}$, \quad
    Boxing Chen $^{1}$, \quad 
    Marzieh S. Tahaei $^{1}$, \quad
}
\affiliations{
    $^{1}$Huawei Noah’s Ark Lab \hspace{0.3cm}
    $^{2}$University of Waterloo \hspace{0.3cm} \\
    aref.jafari@uwaterloo.ca, \hspace{0.3cm} 
    nicole.fan1@h-partners.com 

}

\usepackage{bibentry}

\begin{document}
\pagestyle{plain}
\maketitle

\begin{abstract}

Transformers have demonstrated strong performance across a wide range of sequence modeling tasks, but their quadratic attention complexity limits scalability to long sequences. Linear models such as Mamba and sliding-window attention (SWA) address this by mixing tokens through recurrent or localized operations with fixed-size memory, achieving efficient inference. However, these methods risk degrading performance on long sequences due to their inability to retain detailed information from distant tokens. We propose SCOUT (Segment Compression for Optimized Utility in Transformers), a hybrid architecture that compresses tokens locally within fixed-size segments and applies attention only over these compressed representations. Each token embedding is first enriched via a linear local mixer, Mamba or SWA, that integrates recent context. Then, instead of attending to all previous tokens, each token sparsely attends to a small number of compressed checkpoint tokens that summarize the input history. This design retains much of the expressivity of full attention while substantially reducing the computational and memory cost. By attending to compressed history rather than all previous tokens, SCOUT incurs slightly higher memory than purely linear models, but its growth rate remains sub-quadratic and far more scalable than that of full Transformers. We analyze SCOUT’s computational and memory efficiency and evaluate it empirically on long-context language modeling and reasoning tasks. SCOUT with both Mamba and SWA mixers outperforms strong long-sequence baselines under the same computational budget,  matches full-attention Transformers on language modeling and common-sense reasoning tasks at 400M and 1.3B scales.  Moreover, our SCOUT achieves higher end‑to‑end throughput than state‑of‑the‑art linear models, while delivering comparable results on Long sequence benchmarks. These findings establish SCOUT as a practical, scalable solution for long-range sequence modeling, offering more than 10× savings in compute and memory over full attention.\footnote{The code is released at \textcolor{blue}{https://github.com/ArefJafari/SCOUT}}

\end{abstract}

\begin{figure*}
    \centering
    \includegraphics[clip, trim=0cm 0cm 0cm 2cm, width=0.9\linewidth]{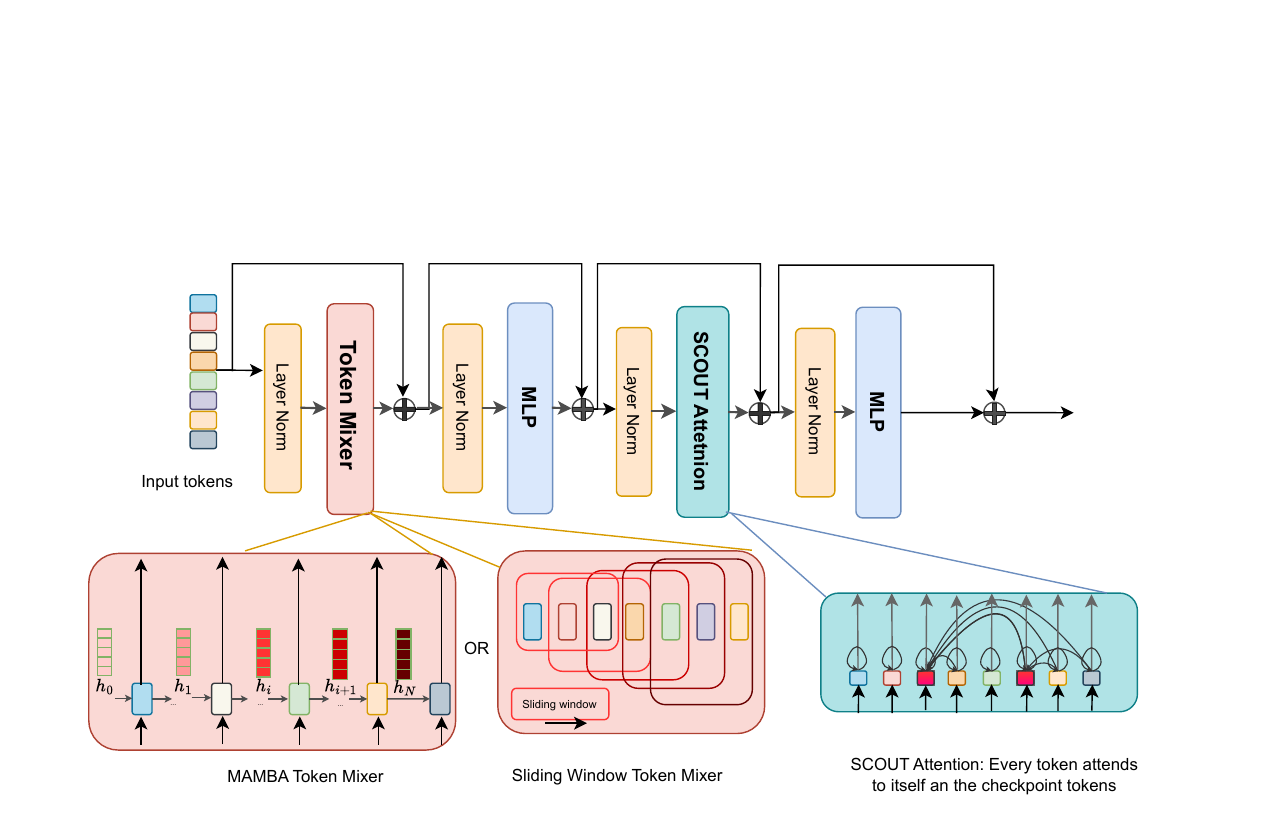}
    \caption{The overview of the SCOUT architecture.}
    \label{fig:main_fig}
\end{figure*}

\section{Introduction}

Transformers~\cite{vaswani2017attention} have emerged as the dominant architecture across natural language processing (NLP), vision, and multimodal tasks, powering state-of-the-art (SOTA) performance in large language models (LLMs) such as GPT-4, Claude, and Gemini~\cite{openai2023gpt4, claude, reid2024gemini}. However, their core mechanism—self-attention—scales quadratically with sequence length in computation and linearly in memory, creating a fundamental bottleneck in long-context scenarios. This limitation restricts the applicability of Transformers in domains requiring long-term reasoning, such as multi-document summarization, scientific literature understanding, and high-resolution time series modeling.

To address this challenge, recent research has explored three major directions. First, linear state-space models (SSMs) aim to replace attention entirely with linear-time alternatives. Models such as Mamba~\cite{gu2023mamba}, GLA~\cite{yang2023gated}, BASED~\cite{arora2024simple}, and DeltaNet~\cite{yang2024parallelizing} compress the input history into a compact recurrent state, enabling efficient processing of long sequences. However, because all historical information must pass through a fixed-size hidden state, these models suffer from fading memory: the influence of earlier tokens diminishes over time, limiting their ability to model long-range dependencies and generalize beyond their training context~\cite{ye2025longmamba, ben2024decimamba}.

Second, hybrid architectures improve expressiveness by combining fast local operations with periodic global context mechanisms. For example, Jamba~\cite{lieber2024jamba} alternates Mamba blocks with full attention layers, while Mamba-in-the-Llama~\cite{wang2024mamba} distills pretrained Transformers into hybrid models composed of Mamba and attention layers. Similarly, some architectures interleave sliding-window attention (SWA) layers with global attention~\cite{ren2024samba}. Although these hybrids offer better scalability and generalization than pure Transformers, they still retain quadratic full-attention layers as bottlenecks, particularly at longer sequence lengths.

Third, sparse attention mechanisms attempt to reduce attention overhead by enforcing structured sparsity in the attention pattern. Notable examples include Longformer~\cite{beltagy2020longformer}, BigBird~\cite{zaheer2020big}, Reformer~\cite{kitaev2020reformer}, and Native Sparse Attention (NSA)~\cite{yuan2025native}. These models restrict token interactions to fixed patterns such as sliding windows, global tokens, or random subsets. However, their sparsity is typically heuristic and input-agnostic, with no guarantee of capturing the most relevant contextual dependencies—especially over long ranges.

In this work, we propose \textbf{SCOUT} (\textit{Segment Compression for Optimized Utility in Transformers}), a novel Transformer layer that combines the efficiency of linear token mixers with the targeted precision of sparse attention (Figure~\ref{fig:main_fig}). SCOUT begins with a linear mixer—either a state-space model like Mamba or a fixed-context attention method like SWA—that processes the full input sequence and integrates recent local context in linear time. These mixers use fixed memory to represent history, which can lead to the fading or loss of earlier information due to the limitations of recurrence (in SSMs) or narrow receptive fields (in SWA).

To mitigate this, SCOUT introduces \textit{checkpoint tokens}: compressed memory slots extracted every $k$ steps that summarize broader segments of past tokens. Each token attends only to itself and a small set of preceding checkpoints, without interacting with other tokens in its own segment. This design preserves the locality captured by the token mixer while selectively reinforcing long-range dependencies through a lightweight sparse attention mechanism.

SCOUT avoids full attention layers entirely, improving upon hybrids and sparse transformers that rely on fixed or computationally expensive mechanisms. It enables each token to access all prior context through a hybrid path: recent tokens via linear mixing, and distant segments via checkpoint attention. This results in sub-quadratic complexity while preserving global contextual coverage.

Importantly, token embeddings produced by the linear mixer already encode historical context. In SSMs, this is achieved through recurrence that propagates information forward in time. SWA captures context within a sliding window. In both cases, SCOUT augments these representations with sparse checkpoint attention to recover global dependencies that may have been attenuated. While recent efforts~\cite{ren2024samba} have explored combining SSMs and SWA, SCOUT provides a more general and compositional solution with superior scalability.

We evaluate SCOUT on a range of long-context language modeling and sequence reasoning tasks. Our results demonstrate that SCOUT matches or exceeds the performance of standard Transformers, significantly reduces memory usage, and improves inference speed. These findings establish SCOUT as a practical and scalable recipe for efficient long-range sequence modeling.

\section{Related Work}

\subsection*{Recurrent State Space Models and Mamba}

Recurrent State Space Models (SSMs) offer a scalable alternative to attention by modeling sequential dependencies through recurrence, achieving linear-time complexity with respect to sequence length. Early models such as S4~\cite{gu2022efficientlymodelinglongsequences} demonstrated that continuous-time SSMs could rival attention mechanisms on long-context benchmarks by encoding global information through structured recurrence.

Mamba~\cite{gu2023mamba} advances this line of work by introducing a \textit{selective SSM} that dynamically controls information flow using input-dependent transition and projection parameters. Unlike traditional SSMs with fixed dynamics, Mamba enables each token to modulate how much past information is retained or forgotten, enhancing its ability to model complex temporal dependencies.

Formally, given an input sequence \(X \in \mathbb{R}^{n \times d}\), Mamba computes hidden representations via the following recurrence:

\[
h_t = A_t\, h_{t-1} + B_t\, x_t, \qquad
y_t = C_t^\top h_t \,,
\]

where \(h_t \in \mathbb{R}^N\) is the hidden state at time \(t\), and the matrices \(A_t \in \mathbb{R}^{N \times N}\), \(B_t \in \mathbb{R}^{N \times d}\), and vector \(C_t \in \mathbb{R}^{N}\) are dynamically computed from the input \(x_t\). These learned input-dependent parameters allow Mamba to implement selective memory updates with minimal overhead.

While Mamba achieves high throughput and strong empirical performance, its reliance on a fixed-size hidden state introduces a limitation common to all recurrent architectures: as the sequence length increases, earlier inputs may be overwritten in the hidden state. This phenomenon can hinder the model’s ability to recover fine-grained information from distant tokens, especially in long-context extrapolation settings~\cite{gu2023mamba, ye2025longmamba}.

\subsection*{Sliding-Window Attention (SWA)}

Sliding-Window Attention (SWA) is a restricted self-attention mechanism that limits each token's receptive field to a fixed-size local window of its recent neighbors. This design reduces the quadratic complexity of full attention to linear complexity in sequence length, making it suitable for processing long sequences~\cite{beltagy2020longformer}.

Formally, given an input sequence \(X \in \mathbb{R}^{n \times d}\), each token \(x_t\) attends only to tokens within a window of size \(w\), resulting in:
\[
\tilde{x}_t = \sum_{j = t-w}^{t} \alpha_{tj} \, x_j, \quad
\alpha_{tj} = \mathrm{softmax}\left( \frac{q_t^\top k_j}{\sqrt{d}} \right),
\]
where \(q_t = x_t W_Q\), \(k_j = x_j W_K\), and \(x_j\) are the input embeddings of the windowed context. The attention weights \(\alpha_{tj}\) are computed over the windowed set \([t-w, t]\), ensuring that attention remains causal and localized.

This locality constraint enables SWA to scale linearly with sequence length and improves cacheability and memory efficiency. However, because each token only attends to a small neighborhood, SWA lacks a global view of the sequence. As a result, it is inherently limited in modeling dependencies beyond the sliding window, especially in tasks that require reasoning over distant. 






\subsection*{Sparse Attention Mechanisms}

Sparse attention models reduce complexity by limiting token interactions via fixed patterns—sliding windows (Longformer~\cite{beltagy2020longformer}), random blocks (BigBird~\cite{zaheer2020big}), LSH (Reformer~\cite{kitaev2020reformer}), or learned masks (SPARSEK~\cite{lou2024sparsek}, NSA~\cite{yuan2025nsa}). However, these heuristic and static sparsity patterns may overlook semantically relevant tokens outside predetermined regions~\cite{nawrot2025sparsefrontier}.

\subsection*{Positioning SCOUT}

SCOUT unifies the strengths of recurrent models and local attention mechanisms while addressing their limitations through a compressed sparse attention design.

Unlike SSMs such as Mamba, which encode history via a single hidden state and suffer from fading memory, SCOUT introduces checkpoint tokens to preserve and retrieve long-range information. This mitigates the loss of distant context with a very slow increment in memory size.

Compared to SWA, which limits each token’s view to a fixed local window, SCOUT augments local token mixing with sparse attention over global checkpoints, enabling long-range dependency modeling without quadratic cost.

SCOUT also improves over sparse attention methods by avoiding fixed or random patterns. Instead, it uses structured, content-aligned checkpoint attention that scales sub-quadratically while remaining expressive.

Through this hybrid design, SCOUT achieves efficient and scalable long-sequence modeling with both local fidelity and global awareness.


\begin{algorithm}[t]
\caption{SCOUT Layer Forward Pass}
\label{alg:scout}
\begin{algorithmic}[1]
\Require Input tokens $X \in \mathbb{R}^{n \times d}$, segment size $k$
\Ensure Output $O \in \mathbb{R}^{n \times d}$
\State $X_1 \gets \mathrm{LN}(X)$ 
\State $X_2 \gets X + \mathrm{LTM}(X_1)$ 
\State $\widetilde{X} \gets X_2 + \mathrm{MLP}(\mathrm{LN}(X_2))$
\State Define $\mathcal{I} = \{k, 2k, \dots, \lfloor n/k \rfloor \cdot k\}$
\State $C \gets \widetilde{X}_{\mathcal{I}, :}$
\State Project: $Q = \widetilde{X}W^Q$, $K = \widetilde{X}W^K$, $V = \widetilde{X}W^V$, $K_C = K_{\mathcal{I}, :}$, $V_C = V_{\mathcal{I}, :}$
\State $A_{\text{comp}} = Q K_C^\top$
\State $D_{\text{self}} = (Q \odot K)\mathbf{1}_d$ 
\State $A = \text{softmax}\left( \frac{[A_{\text{comp}} \; D_{\text{self}}]}{\sqrt{d}} \right)$
\State $[\widetilde{A}_{\text{comp}}, \widetilde{D}_{\text{self}}] \gets A$
\State $O = \widetilde{A}_{\text{comp}} V_C + \widetilde{D}_{\text{self}} \odot V$
\State $Y_1 \gets O + \widetilde{X}$
\State $\Return Y_1 + \text{MLP}(\text{LN}(Y_1))$
\end{algorithmic}
\end{algorithm}

\begin{table*}[t]
\centering
\small
\resizebox{0.95\textwidth}{!}{%
\setlength{\tabcolsep}{5pt}
\begin{tabular}{ll|cc|ccccccccc}
\hline
\hline

\multirow{2}{*}{}  & \textbf{Model} & \textbf{Wiki} & \textbf{LMB} & \textbf{LMB} & \textbf{PIQA} & \textbf{Hella} & \textbf{ARC-c} & \textbf{ARC-e} & \textbf{MMLU} & \textbf{CSQA} & \textbf{Avg.} \\
               & & ppl $\downarrow$ & ppl $\downarrow$ & acc $\uparrow$ & acc $\uparrow$ & acc$_n$ $\uparrow$ & acc$_n$ $\uparrow$ & acc $\uparrow$ & acc $\uparrow$ & acc $\uparrow$ & acc
               $\uparrow$
               \\               
\hline

\multirow{6}{*}{\rotatebox{90}{$\approx$400\,M}}&
LLaMA               & 29.03 & 45.91 & 31.36 & 66.16 & 38.05 & 26.71 & 56.94 & 23.27 & 19.66 & 37.45 \\
& LLaMA-SWA           & 28.31 & 44.44 & 30.89 & 66.54 & 38.50 & 27.82 & 57.79 & 25.25 & 19.49 & 38.04 \\
& Mamba               & 56.91 & 292.26 & 13.45 & 61.15 & 30.64 & 24.40 & 46.59 & 23.06 & 19.66 & 31.28 \\
& GLA                 & 32.42 & 48.57 & 28.66 & 65.67 & 37.21 & 24.32 & 55.22 & 23.30 & 20.64 & 36.43 \\
& DeltaNet            & 31.15 & 51.13 & 28.37 & 65.67 & 37.02 & 26.62 & 55.05 & 23.37 & 19.74 & 36.55 \\
& SCOUT-SWA           & 28.35 & 42.69 & 32.51 & 67.19 & 39.18 & 28.92 & 55.81 & 23.62 & 19.90 & 38.16 \\

& SCOUT-Mamba         & 29.03 & 45.91 & 31.36 & 66.26 & 38.75 & 27.13 & 57.41 & 24.53 & 19.16 & 37.80 \\
\hline
\hline
\multirow{6}{*}{\rotatebox{90}{$\approx$1.3\,B}}&
LLaMA               & 17.22 & 14.61 & 44.98 & 71.71 & 55.30 & 36.95 & 70.58 & 25.63 & 19.33 & 46.35 \\
& LLaMA-SWA           & 16.88 & 14.56 & 45.60 & 72.47 & 55.97 & 37.97 & 70.71 & 25.33 & 21.38 & 47.06 \\
& Mamba               & 17.30 & 15.43 & 44.52 & 71.27 & 50.52 & 28.07 & 57.28 & 24.51 & 18.84 & 42.14 \\

& SCOUT-SWA           & 17.49 & 14.61 & 45.66 & 72.42 & 55.35 & 38.40 & 71.00 & 25.40 & 20.80 & 47.00 \\
& SCOUT-Mamba         & 18.04 & 14.91 & 44.30 & 71.82 & 55.08 & 38.23 & 71.00 & 24.33 & 20.31 & 46.44 \\
\hline
\hline
\end{tabular}}
\caption{Performance comparison on language modeling and reasoning tasks for models around 400M and 1.3B parameter scale under ... TFlops and 6 TFLOPs respectively. Last column reports the average across reasoning scores.}

\label{tab:main_results}
\end{table*}

\section{Methodology}

SCOUT is a transformer layer that achieves sub-quadratic attention complexity by combining efficient linear token mixing with sparse, structured attention over compressed memory. Unlike sparse attention models that heuristically attend to parts of the sequence, SCOUT ensures all input information is incorporated via a hybrid mechanism: local recurrence that captures recent tokens, while attention over compressed checkpoints reinforces long-range memory.

\subsection{Architecture Overview}
Each SCOUT layer consists of three components:
\begin{enumerate}
    \item A Mamba or SWA token mixer that encodes the current token with the entire past sequence via a recurrent state.
    \item A standard position-wise feedforward network (MLP).
    \item A sparse attention mechanism over compressed checkpoint tokens to recover long-range dependencies.
    \item An other standard position-wise feedforward network (MLP).
\end{enumerate}
This structure avoids full attention layers and replaces them with compressed, memory-efficient sparse routing.

\subsection{Token Mixing}

Let \( X \in \mathbb{R}^{n \times d} \) denote the input sequence of \( n \) tokens, each with hidden size \( d \). SCOUT begins with a linear token mixer (LTM) that integrates recent contextual information into each token. The choice of token mixer depends on the variant: SCOUT-Mamba uses a recurrent state-space model (Mamba), while SCOUT-SWA employs causal sliding-window attention.

The token mixing stage proceeds as follows:

\begin{align*}
X_1 &= \mathrm{LN}(X) \\
X_2 &= X + \mathrm{LTM}(X_1) \\
\widetilde{X} &= X_2 + \mathrm{MLP}(\mathrm{LN}(X_2))
\end{align*}

Here, \( \mathrm{LN} \) denotes layer normalization, and the MLP is a feedforward network. The residual connections improve training stability.

In SCOUT-Mamba, the LTM captures all previous tokens dependencies via a recurrent hidden state, but earlier tokens may fade due to limited memory capacity. In SCOUT-SWA, each token attends to a fixed-size local window, restricting its context to nearby tokens. In both cases, \( \widetilde{X} \in \mathbb{R}^{n \times d} \) encodes the current token along with recent context, but long-range dependencies may weaken—an issue addressed in the next stage by sparse checkpoint attention. Note that this stage runs in linear time $O(n)$ with respect to the sequence length.

\subsection{Checkpoint Compression}
To recover potentially faded memory from the previous step, we introduce a checkpointing mechanism that captures long-range context at regular intervals across the input sequence. Let the input sequence have length $n$, and let $k$ be a fixed interval defining the checkpoint spacing. A set of checkpoint indices $I$ is defined as:
\[
\mathcal{I} = \{k, 2k, 3k, \dots, \lfloor n/k \rfloor \cdot k\}
\]
This results in approximately $n/k$ equally spaced checkpoints over the sequence. A compressed memory matrix $C$ is then constructed by selecting the hidden representations (i.e., rows of the intermediate output  $\widetilde{X}$ at these checkpoint indices):
\[
C = \widetilde{X}_{\mathcal{I}, :} \in \mathbb{R}^{(n/k) \times d}
\]

\subsection{Checkpoint Attention Mechanism (SCOUT Attention)}

To recover long-range dependencies lost in the token mixer stage, SCOUT employs a lightweight and efficient attention mechanism. Rather than attending to all previous tokens (as in full self-attention), each token attends only to:
\begin{itemize}
    \item Itself, via a diagonal self-attention term.
    \item A small number of compressed checkpoint tokens, via sparse attention over memory.
\end{itemize}
This design enables SCOUT to access global context with sub-quadratic cost while maintaining linear access to recent history through the token mixer.

Let \( Q = \widetilde{X} W^Q \), \( K = \widetilde{X} W^K \), and \( V = \widetilde{X} W^V \), where \( W^Q, W^K, W^V \in \mathbb{R}^{d \times d} \) are learnable projection matrices. Let \( C \in \mathbb{R}^{(n/k) \times d} \) denote the matrix of compressed checkpoint tokens extracted from the token-mixed representation \( \widetilde{X} \). We define the projected checkpoint keys and values as:
\[
K_C = C W^K, \quad V_C = C W^V
\]

\paragraph{Compressed Sparse Attention.}
Each token computes attention scores over the checkpoint tokens:
\[
A_{\text{comp}} = Q K_C^\top + \mathcal{M} \in \mathbb{R}^{n \times (n/k)}
\]
where \( \mathcal{M} \) is a causal mask to prevent attending to future checkpoints. This operation has total complexity $O(n^2/k)$

\paragraph{Diagonal Self-Attention.}
In parallel, each token attends to itself through a diagonal term computed as:
\[
D_{\text{self}} = (Q \odot K) \mathbf{1}_d
\]
where \( \odot \) denotes the Hadamard product and \( \mathbf{1}_d \in \mathbb{R}^{d \times 1} \) is a vector of ones. This yields a self-attention score vector in \( \mathbb{R}^{n} \) where \( D_{\text{self}}[t] = Q_t \cdot K_t \). This operation has total complexity \( \mathcal{O}(n) \).

\paragraph{Attention Aggregation.}
We concatenate the compressed and diagonal scores and apply softmax normalization:
\[
A = \text{softmax}\left( \frac{[A_{\text{comp}} \;\; D_{\text{self}}]}{\sqrt{d}} \right)
\]
Let \( A = [\widetilde{A}_{\text{comp}} \;\; \widetilde{D}_{\text{self}}] \in \mathbb{R}^{n \times (n/k + 1)} \), where the final attention output is:
\[
O = \widetilde{A}_{\text{comp}} V_C + \widetilde{D}_{\text{self}} \odot V
\]
Here, \( \widetilde{D}_{\text{self}} \odot V \) represents row-wise scaling of each token’s value vector \( V_t \) by its corresponding self-attention weight \( \widetilde{D}_{\text{self}}[t] \).

This hybrid attention mechanism enables each token to access:
\begin{itemize}
    \item \textbf{Recent context}, via local mixing in linear token mixer (e.g., Mamba or SWA),
    \item \textbf{Global context}, via sparse attention over compressed checkpoints,
\end{itemize}
Together, these components enable SCOUT to model long sequences efficiently with a computational complexity significantly lower than full attention while preserving long-range contextual capacity.

\subsection{Feedforward Network and Residual Connections}
The output $O$ is passed through a standard transformer MLP:
\begin{align*}
Y_1 &= O + \widetilde{X}\\
Y &= Y_1 + \text{MLP}(\text{LN}(Y_1)) \\
\end{align*}
For a detailed summary of the SCOUT architecture and its forward computation, please refer to Algorithm \ref{alg:scout}

\subsection{Efficiency Analysis}

SCOUT significantly reduces memory and compute requirements by leveraging compressed attention over a small set of checkpoint tokens. Instead of caching full key and value matrices of size \( \mathbb{R}^{n \times d} \), SCOUT only stores the projected checkpoint tokens:
\[
K_C, V_C \in \mathbb{R}^{(n/k) \times d}
\]
This reduces memory complexity from \( \mathcal{O}(n) \) to \( \mathcal{O}(n/k) \).

In terms of computation, SCOUT attention performs dot products between all query tokens and only the checkpoint keys, leading to a total cost of:
\[
\mathcal{O}(n \cdot (n/k)) = \mathcal{O}(n^2 / k)
\]
Thus, SCOUT reduces both memory and computational overhead by a factor of \( k \) compared to full attention. In our experiments, we set \( k = 10 \) for default runs and explore up to \( k = 50 \) in ablations (see Appendix). These settings yield 10× to 50× reductions in memory and compute, with almost no degradation in accuracy.

Importantly, SCOUT demonstrated robust performance in our ablation studies on language modeling tasks, even as \( k \) increased, highlighting its robustness to checkpoint sparsity. As \( k \to n \), SCOUT approaches linear complexity while still capturing global dependencies through its sparse attention mechanism.

\begin{figure*}[t]
    \centering
    \includegraphics[width=\linewidth]{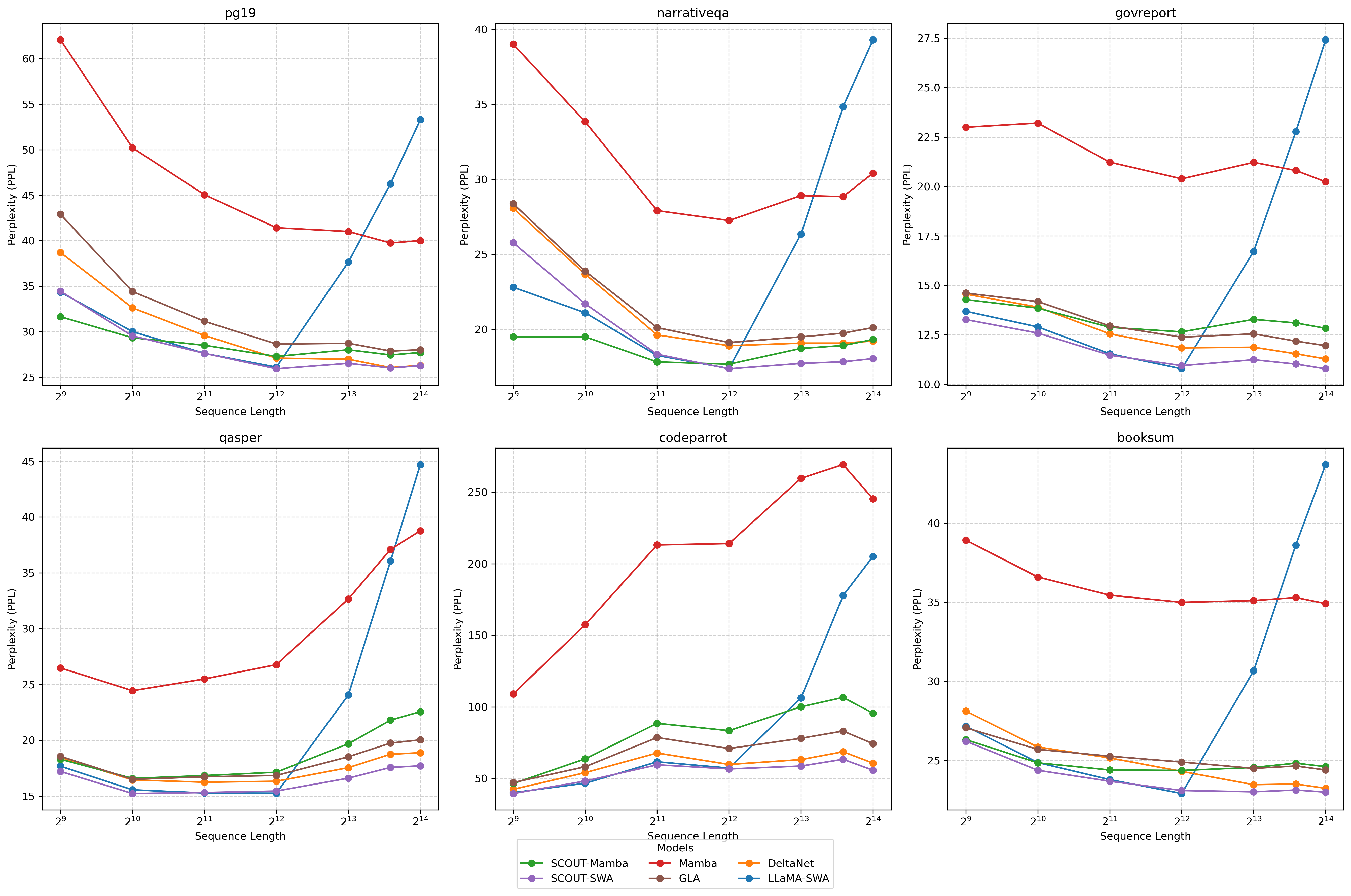}
    \caption{Perplexity comparison on six long-context benchmarks: PG-19, BookSum, NarrativeQA, GovReport, Qasper, and CodeParrot, using 400M scale models. SCOUT-SWA achieves the lowest perplexity across all sequence lengths, outperforming all baselines. SCOUT-Mamba shows significant improvement over its Mamba backbone and is competitive with other linear models like GLA and RetNet.}
    \label{fig:ppl_400M}
\end{figure*}

\begin{table*}[t]
\centering
\small
\setlength{\tabcolsep}{4pt}
\begin{tabular}{l|
cc|ccc|ccc|ccc|cc|c}
\hline
\textbf{Model} &
\multicolumn{2}{c|}{\textbf{Single-Doc QA}} &
\multicolumn{3}{c|}{\textbf{Multi-Doc QA}} &
\multicolumn{3}{c|}{\textbf{Summarization}} &
\multicolumn{3}{c|}{\textbf{Retrieval}} &
\multicolumn{2}{c|}{\textbf{Code}} &
\textbf{Avg.} \\
& TQA & TRC & 2WM & HPQA & QAS & GvR & MNS & SMS & MFQ & PCT & PRet & LCC & RBP & \\
\hline
LLaMA         & 15.12 & 3.44 & 11.88 & 12.35 & 16.06 & 4.27 & 6.40 & 7.37 & 15.32 & 1.84 & 2.36 & 6.43 & 4.17 & 8.23 \\
LLaMA-SWA     & 19.52 & 2.33 & 21.01 & 17.63 & 19.04 & 3.36 & 5.43 & 4.70 & 16.91 & 0.94 & 5.72 & 9.75 & 9.25 & 10.43 \\
Mamba         & 29.31 & 3.06 & 33.36 & 28.04 & 23.80 & 16.80 & 14.14 & 6.67 & 29.01 & 2.53 & 3.55 & 28.71 & 24.37 & 18.72 \\
SCOUT-Mamba   & 32.79 & 10.94 & 35.93 & 35.29 & 31.28 & 8.33 & 7.06 & 10.19 & 37.02 & 2.03 & 4.11 & 25.23 & 25.85 & 20.47 \\
SCOUT-SWA     & 36.81 & 14.89 & 37.62 & 29.87 & 28.41 & 10.82 & 10.09 & 15.11 & 37.40 & 0.95 & 4.33 & 9.53 & 9.99 & 18.91 \\
\hline
\end{tabular}
\caption{
Accuracy on 13 tasks from LongBench\_e (Bai et al., 2023): TriviaQA, TREC, 2WikiMultiQA, HotpotQA, Qasper, GovReport, MultiNews, Samsum, MultiFieldQA, PassageCount, PassageRetrieval, LCC, and RepoBench-P, following the original evaluation order.}
\label{tab:longbench}
\end{table*}

\begin{figure*}[t]
    \centering
    \includegraphics[width=0.8\linewidth]{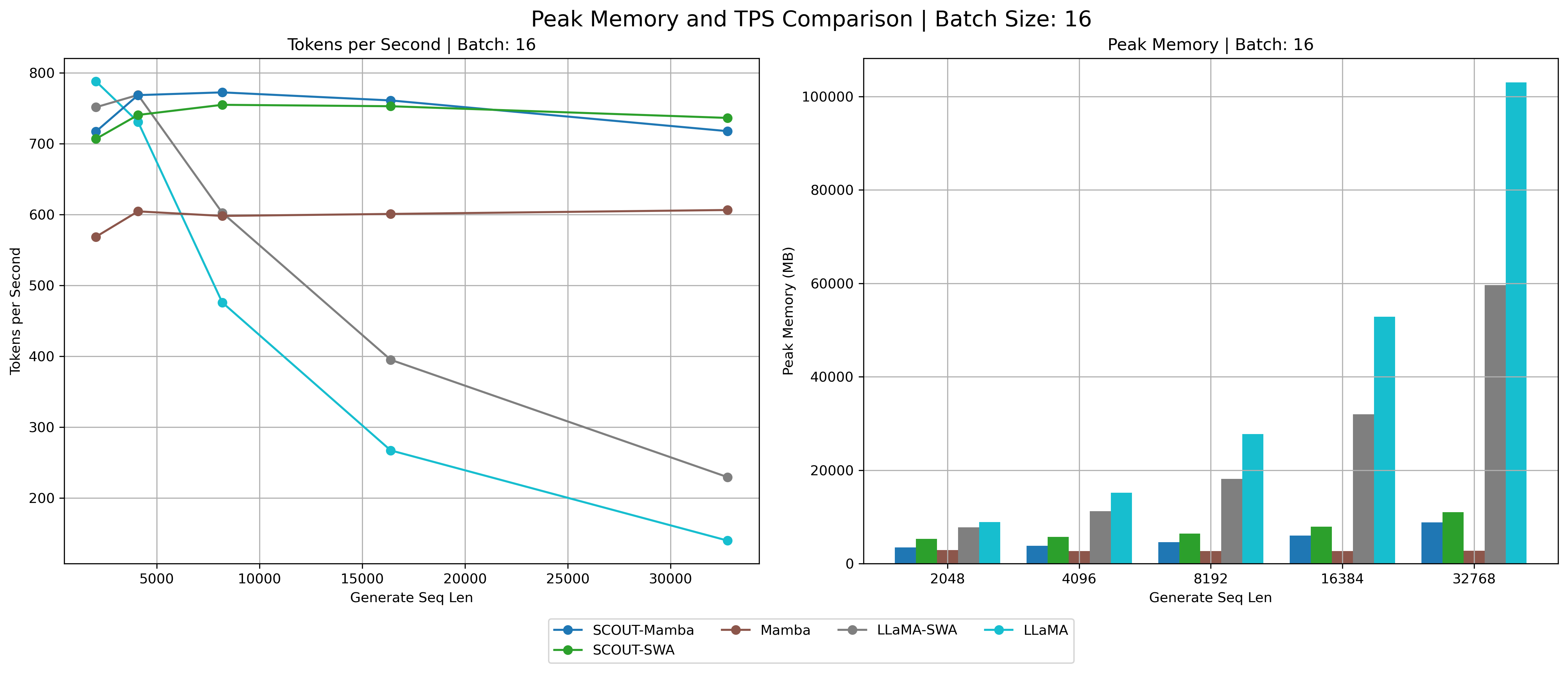}
    \caption{
    \textbf{Generation throughput and peak memory usage} for models at 1B scale, evaluated with a batch size of 16. 
    Left: Tokens per second (TPS) across generation lengths from 2K to 32K. 
    Right: Peak memory usage during generation.
    }
    \label{fig:latency}
\end{figure*}



\section{Experiments}
\subsection{Experimental Setup.}
\paragraph{Training details}We conduct a systematic evaluation of recent high-performing sequence modeling architectures across two model scales: 400M and 1B parameters.  
The 400M-scale models are trained on 15B tokens, while the 1B-scale models are trained on 100B tokens, both drawn from the FineWeb-Edu corpus.  
All models use the \textsc{LLaMA2} tokenizer with a shared vocabulary of 32,000 tokens to ensure consistency across comparisons.  
Training is performed using the AdamW optimizer with a peak learning rate of $3 \times 10^{-4}$, weight decay of 0.1, gradient clipping of 1.0, and a cosine learning rate schedule including a linear warm-up phase.  
Batch size is set to 0.5M tokens for the 400M models and 2M tokens for the 1B models.  
The training sequence length is fixed at 4K for 400M models and 2K for 1B models.  
All models are trained under matched compute budgets to ensure a fair comparison of architectural trade-offs.

\paragraph{FLOPs based comparison.}To ensure a fair comparison across models with fundamentally different architectural designs, we align models based on their FLOPs (floating point operations) rather than parameter count. Matching solely on the number of parameters can be misleading, as different architectures (e.g., recurrent, sparse, or attention-based) utilize parameters with varying computational and representational efficiency. Therefore, for Table~\ref{tab:main_results}, we match FLOPs at a fixed input length of 2K tokens. Specifically, models in the 400M-scale group are calibrated to operate within approximately 4 TFLOP per forward pass, while models in the 1.3B-scale group are calibrated around 6 TFLOPs. At the 1.3B scale, LLaMA has 1.36B parameters, LLaMA-SWA has 1.42B, SCOUT-SWA has 1.57B, and SCOUT-Mamba reaches 1.59B. For the 400M scale, LLaMA contains 340M parameters, LLaMA-SWA has 370M, SCOUT-SWA has 470M, and SCOUT-Mamba has 430M. For comparison, the baseline Mamba model has 550M parameters, while GLA and DeltaNet are each configured with 340M parameters.
The exact hidden dimensions for each model variant are chosen to meet this constraint and are detailed in Appendix. Notably, this setup enables SCOUT variants to reduce FLOPs further when evaluated at longer sequence lengths, due to their sub-quadratic attention design. This property is further analyzed in Section Results, Latency and Memory Efficiency, where we demonstrate that SCOUT maintains high throughput and stable memory usage as sequence length increases, in contrast to standard attention-based architectures whose cost grows quadratically.

\subsection{Results}
\paragraph{Language Modeling} In 
Table~\ref{tab:main_results}, we present language modeling perplexity and zero-shot accuracy for models at 400M and 1.3B scales. 
At the 400M scale, our SCOUT-SWA achieves the best overall performance, with the lowest perplexity on Wiki and LMB and the highest average accuracy across reasoning benchmarks. SCOUT-Mamba also improves significantly over the base Mamba, reducing perplexity and improving accuracy across the board. 
At the 1.3B scale, SCOUT-SWA remains competitive, closely matching LLaMA-SWA in reasoning accuracy (47.00 vs. 47.06) while outperforming LLaMA and Mamba.  Meanwhile, SCOUT-Mamba again outperforms the linear Mamba baseline by a wide margin in both perplexity and accuracy  even outperforming the full Transformer (LLaMA). These results validate SCOUT’s ability to enhance linear token mixers with sparse compressed attention, enabling strong modeling performance under strict FLOPs budgets.

\paragraph{Length Extrapolation.} In Figure~\ref{fig:ppl_400M}, we evaluate model performance on sequences up to 16K tokens across six long-context language modeling benchmarks, includingPG-19, BookSum, NarrativeQA, GovReport, Qasper, and CodeParrot. SCOUT-SWA consistently achieves the lowest perplexity across all sequence lengths, outperforming all other baselines, including GLA, DeltaNet, Mamba, and LLaMA-SWA; and demonstrating strong generalization beyond the training horizon of 4K tokens. In contrast, LLaMA-SWA fails to extrapolate, with perplexity rising sharply past its training limit. SCOUT-Mamba also significantly improves over its backbone Mamba, and delivers performance comparable to other linear-time baselines like GLA and RetNet, highlighting its effectiveness in extending the reach of recurrent models.

\paragraph{Long Context Understanding} As shown in Table~\ref{tab:longbench}, we evaluate models on 13 long-context tasks from LongBench\_extension, spanning QA, summarization, retrieval, and code understanding. SCOUT-Mamba achieves the highest average score, demonstrating strong retrieval and code reasoning abilities with state tracking advantages over Mamba. 
SCOUT-Mamba achieves the highest overall average score, with strong performance on multi-document QA and code understanding tasks, underscoring its ability to model long-range dependencies through state tracking. SCOUT-SWA performs best on single-document QA and summarization tasks, and ranks second overall, remaining highly competitive. Both SCOUT variants consistently outperform their respective backbones (Mamba and LLaMA-SWA) as well as the full LLaMA baseline, demonstrating the effectiveness of SCOUT’s sparse compressed attention for long-context reasoning.

\paragraph{Latency and Memory Efficiency}
Figure~\ref{fig:latency} shows the generation throughput (tokens per second) and peak memory usage across sequence lengths ranging from 2K to 32K for models at the 1.3B scale. SCOUT-SWA and SCOUT-Mamba consistently achieve the highest throughput across all settings, outperforming even the linear baseline Mamba in speed, while maintaining stable performance as sequence length increases. Note that this speed advantage over mamba arises because, under a matched FLOPs budget at 2K, SCOUT uses fewer layers than the standard Mamba architecture, resulting in faster inference.
This contrasts with full-attention models such as LLaMA and hybrid models like LLaMA-SWA, whose throughput declines sharply beyond the 4K context window due to their quadratic attention layers. In terms of memory usage, SCOUT’s sub-quadratic attention results in slightly higher consumption than fully linear Mamba, but remains significantly more efficient than hybrid models like LLaMA-SWA and full attention models like LLaMA, whose memory usage grows steeply with sequence length.


\section{Conclusion}

We introduced SCOUT, a sub-quadratic transformer framework that augments efficient token mixers with sparse attention over compressed checkpoints. By decoupling local token mixing from global context retrieval, SCOUT enables scalable and expressive sequence modeling without relying on quadratic full attention. We instantiated SCOUT with both Mamba and sliding window attention (SWA), demonstrating that this hybrid design can recover long-range dependencies while maintaining high throughput and memory efficiency. 

Through extensive experiments on language modeling, long-context benchmarks, and structured retrieval tasks, we showed that SCOUT consistently improves over its linear baselines,  boosting performance over linenar and hybrid architectures. These results establish SCOUT as a practical and versatile recipe for long-context modeling, combining the efficiency of linear models with the representational strength of compressed global memory. We hope this design encourages further research into structured, content-aware compression as a foundation for scaling transformers to ever longer sequences.


\bibliography{SCOUT_aaai2026}

\clearpage
\appendix

\begin{table}[H]
\centering
\small
\setlength{\tabcolsep}{5pt}
\begin{tabular}{l|cc|ccccccccc}
\hline
\textbf{Model} & \textbf{Wiki} & \textbf{LMB} & \textbf{LMB} & \textbf{PIQA} & \textbf{Hella} & \textbf{ARC-c} & \textbf{ARC-e} & \textbf{MMLU} & \textbf{CSQA} & \textbf{Avg.} \\
               & ppl $\downarrow$ & ppl $\downarrow$ & acc $\uparrow$ & acc $\uparrow$ & acc$_n$ $\uparrow$ & acc$_n$ $\uparrow$ & acc $\uparrow$ & acc $\uparrow$ & acc $\uparrow$ & \\
\hline
SCOUT-SWA (k=10, s=1024) & 28.35 & 42.69 & 32.51 & 67.19 & 39.18 & 28.92 & 55.81 & 23.62 & 19.90 & 38.16 \\
SCOUT-SWA (k=20, s=1024) & 28.52 & 43.28 & 33.17 & 66.00 & 39.05 & 27.39 & 57.07 & 24.58 & 21.13 & 38.34 \\
SCOUT-SWA (k=50, s=1024) & 28.14 & 40.61 & 33.40 & 65.83 & 39.36 & 27.47 & 57.70 & 24.11 & 19.82 & 38.24 \\
\hline
\end{tabular}
\caption{Ablation on checkpoint interval $k$ under fixed sliding window size $s=1024$ in 400M scale models.}
\label{tab:k-ablation}
\end{table}

\begin{table}[H]
\centering
\small
\setlength{\tabcolsep}{5pt}
\begin{tabular}{l|cc|ccccccccc}
\hline
\textbf{Model} & \textbf{Wiki} & \textbf{LMB} & \textbf{LMB} & \textbf{PIQA} & \textbf{Hella} & \textbf{ARC-c} & \textbf{ARC-e} & \textbf{MMLU} & \textbf{CSQA} & \textbf{Avg.} \\
               & ppl $\downarrow$ & ppl $\downarrow$ & acc $\uparrow$ & acc $\uparrow$ & acc$_n$ $\uparrow$ & acc$_n$ $\uparrow$ & acc $\uparrow$ & acc $\uparrow$ & acc $\uparrow$ & \\
\hline
SCOUT-SWA (k=10, s=512)  & 29.15 & 42.83 & 31.94 & 66.21 & 40.11 & 27.05 & 57.15 & 24.01 & 20.39 & 38.12 \\
SCOUT-SWA (k=10, s=1024) & 28.35 & 42.69 & 32.51 & 67.19 & 39.18 & 28.92 & 55.81 & 23.62 & 19.90 & 38.16 \\
\hline
\end{tabular}
\caption{Ablation on sliding window size $s$ under fixed checkpoint interval $k=10$ in 400M scale models.}
\label{tab:s-ablation}
\end{table}

\begin{table}[H]
\centering
\small
\setlength{\tabcolsep}{5pt}
\begin{tabular}{l|cc|ccccccccc}
\hline
\textbf{Model} & \textbf{Wiki} & \textbf{LMB} & \textbf{LMB} & \textbf{PIQA} & \textbf{Hella} & \textbf{ARC-c} & \textbf{ARC-e} & \textbf{MMLU} & \textbf{CSQA} & \textbf{Avg.} \\
               & ppl $\downarrow$ & ppl $\downarrow$ & acc $\uparrow$ & acc $\uparrow$ & acc$_n$ $\uparrow$ & acc$_n$ $\uparrow$ & acc $\uparrow$ & acc $\uparrow$ & acc $\uparrow$ & \\
\hline
SCOUT-SWA (w/ MLP)        & 28.35 & 42.69 & 32.51 & 67.19 & 39.18 & 28.92 & 55.81 & 23.62 & 19.90 & 38.16 \\
SCOUT-SWA (w/o MLP)       & 28.40 & 45.78 & 31.40 & 66.05 & 39.05 & 28.58 & 57.32 & 23.81 & 20.97 & 38.17 \\
SCOUT-Mamba (w/ MLP)      & 31.32 & 46.14 & 28.66 & 66.26 & 38.75 & 27.13 & 57.41 & 24.53 & 19.16 & 37.41 \\
SCOUT-Mamba (w/o MLP)     & 31.93 & 48.12 & 28.35 & 66.05 & 38.11 & 26.02 & 56.14 & 23.40 & 18.34 & 36.63 \\
\hline
\end{tabular}
\caption{Ablation on the presence of intermediate MLP layers in SCOUT variants in 400M scale models.}
\label{tab:mlp-ablation}
\end{table}

\section{Appendix}
\subsection{A1. Effect of Checkpoint Interval ($k$) and Window Size ($s$)}

We perform ablation studies on SCOUT-SWA at the 400M model scale using a 15B-token subset of the FineWeb-Edu dataset. Two key hyperparameters are varied: the checkpoint interval $k$, which controls how frequently checkpoint tokens are stored, and the sliding window size $s$, which determines the local receptive field of each token. A larger $k$ leads to sparser memory updates, while a larger $s$ expands the range of accessible past context. As shown in Tables~\ref{tab:k-ablation} and~\ref{tab:s-ablation}, performance remains relatively stable across different configurations, indicating that SCOUT-SWA is robust to these hyperparameters in the low-resource setting.

\vspace{1em}
\subsection{A2. Effect of Intermediate MLP Layer}

We also study the impact of including an intermediate MLP layer between the linear token mixer (LTM block) and the SCOUT attention block. Table~\ref{tab:mlp-ablation} compares variants with and without this MLP. Removing the MLP leads to a consistent decline in performance across both SCOUT-SWA and SCOUT-Mamba models. These results highlight the importance of nonlinear transformations for improving expressiveness, and justify the use of MLP-equipped variants in our main experiments.


\end{document}